\documentclass[runningheads]{llncs}
\usepackage{graphicx}
\usepackage{dsfont}         
\usepackage{amsmath,amsfonts,bm} 
\usepackage{nicefrac}       
\usepackage{subcaption} 
\usepackage{algorithm,algorithmic} 

\def\rx{{\textnormal{x}}} 
\def\rr{{\textnormal{r}}} 
\def\rl{{\textnormal{l}}} 
\def\1{\bm{1}} 

\def\mT{{\bm{T}}} 
\def\emT{{T}} 
\def\mD{{\bm{D}}} 
\def\emD{{D}} 
\def\mU{{\bm{U}}} 
\def\mS{{\bm{S}}} 
\def\mV{{\bm{V}}} 
\def\mC{{\bm{C}}} 
\def\mL{{\bm{L}}} 

\def\mM{{\bm{M}}} 
\def\mO{{\bm{O}}} 
\def\mV{{\bm{V}}} 
\def\mlambda{{\bm{\lambda}}} 

\newcommand{\R}{\mathbb{R}} 
\newcommand{\N}{\mathbb{N}} 
\newcommand{\pdata}{p_{\rm{data}}}

\begin{document}
\title{Structured (De)composable Representations Trained with Neural Networks}
%
%
\author{Graham Spinks\inst{1}\orcidID{0000-0002-5490-4879} \and
Marie-Francine Moens\inst{1}\orcidID{0000-0002-3732-9323} }
%
\authorrunning{G. Spinks and M-F. Moens}
%
\institute{KU Leuven, Department of Computer Science, Belgium 
\email{\{graham.spinks,sien.moens\}@cs.kuleuven.be}\\}
\maketitle              
\begin{abstract}

The paper proposes a novel technique for representing templates and instances of concept classes. A template representation refers to the generic representation that captures the characteristics of an entire class. The proposed technique uses end-to-end deep learning to learn structured and composable representations from input images and discrete labels.
The obtained representations are based on distance estimates between the distributions given by the class label and those given by contextual information, which are modeled as environments. We prove that the representations have a clear structure allowing to decompose the representation into factors that represent classes and environments. We evaluate our novel technique on classification and retrieval tasks involving different modalities (visual and language data).

\end{abstract}

\section{Introduction}

\begin{figure}[ht]
	\begin{center}
		\includegraphics[width=0.99\linewidth]{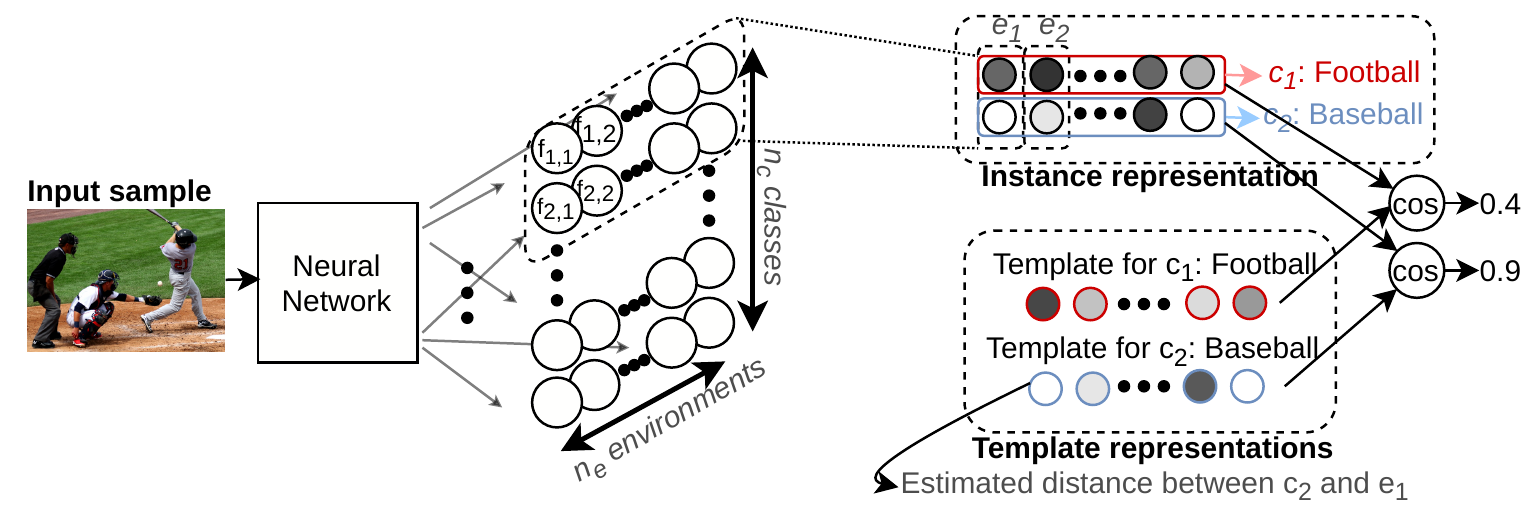}
	\end{center}
	\caption[Template explanation]{The last layer of a convolutional neural network is replaced with fully-connected layers that map to $n_c \times n_e$ outputs $f_{i,j}$ that are used to create instance representations that are interpretable along contextual dimensions, which we call `environments'. By computing the cosine similarity, rows are compared to corresponding class representations, which we refer to as `templates'.}
	\label{fig:templates}
\end{figure}

We propose a novel technique for representing templates and instances of concept classes that is agnostic with regard to the underlying deep learning model. Starting from raw input images, representations are learned in a classification task where the cross-entropy classification layer is replaced by a fully connected layer that is used to estimate a bounded approximation of the distance between each class distribution and a set of contextual distributions that we call `environments'. By defining randomized environments, the goal is to capture common sense knowledge about how classes relate to a range of differentiating contexts, and to increase the probability of encountering distinctive diagnostic features.
This idea loosely resembles how human long-term memory might achieve retrieval \cite{nairne2002myth} as well as how contextual knowledge is used for semantic encoding \cite{murdock1982theory}. Our experiments confirm the value of such an approach.

In this paper, classes correspond to (visual) object labels, and environments correspond to combinations of contextual labels given by either object labels or image caption keywords. 
Representations for individual inputs, which we call `instance representations', form a 2D matrix with rows corresponding to classes and columns corresponding to environments, where each element is an indication of how much the instance resembles the corresponding class versus environment. The parameters for each environment are 
defined once at start by uniformly selecting a randomly chosen number of labels from the power set of all available contextual labels. The class representation, which we refer to as `template', has the form of a template vector. It contains the average distance estimates between the distribution of a class and the distributions of the respective environments. By computing the cosine similarity between between the instance representation and all templates, class membership can be determined efficiently (Fig. \ref{fig:templates}). 

Template and 
instance representations are interpretable as they have a fixed structure comprised of distance estimates. 
This structure is reminiscent of traditional language processing matrix representations and enables operations that operate along matrix dimensions. We demonstrate this with a Singular Value Decomposition (SVD) 
which yields components that determine the values along the rows (classes) and columns (environments) respectively. Those components can then be altered to modify the information content, upon which a new representation can be reconstructed. 
The proposed representations are evaluated in four settings: 
(1) Multi-label image classification, \textit{i.e.}, object recognition with multiple objects per image; 
(2) Image retrieval where we query images that look like existing images but contain altered class labels; (3) Single-label image classification on pre-trained instance representations for a previously unseen label; (4) Rank estimation with regard to compression of the representations.

\textbf{Contributions} (1) We propose a new 
deep learning technique 
to create structured representations from images, entity classes and their contextual information (environments) based on distance estimates. 
(2) This leads to template representations that generalize well, as successfully evaluated in a classification task. 
(3) The obtained representations are interpretable as distances between a class and its environment. They are composable in the sense that they can be modified to reflect different class membership 
as shown in a retrieval task.

\section{Related work}

We shortly discuss related work for different aspects of our research.

\textbf{Representing entities with respect to context.}

In language applications, structured matrices (\textit{e.g}, document-term matrices) have been used for a long time. Such matrices can be decomposed with SVD or non-negative matrix factorization. Low-rank approximations are found with methods like latent semantic indexing. Typical applications are clustering, classification, retrieval, etc. with the benefit that outcomes can usually be interpreted with respect to the contextual information. Contrary to our work, earlier methods build representations purely from labels and don't take 
deep neural network-based features into account.
More recently \cite{singh2019context} create an unsupervised sentence representation where each entity is a probability distribution based on co-occurrence of words.   

\textbf{Distances to represent features.}
The Earth Mover's Distance (EMD) 
also known as Wasserstein distance, is a useful metric based on the optimal transport problem to measure the distance between distributions. \cite{kusner2015word} use a similar idea to define the Word Mover's Distance (WMD) that measures the minimal amount of effort to move Word2Vec-based word embeddings from one document to another. The authors use a matrix representation that expresses the distance between words in respective documents. They note the structure is interpretable and performs well on text-based classification tasks.

\textbf{Random features.}
The Word Mover's Embedding \cite{wu2018word} is an unsupervised feature representation for documents, created by concatenating WMD estimates that are computed with respect to arbitrarily chosen feature maps. The authors calculate an approximation of the distance between a pair of documents with the use of a kernel over the feature map. The building blocks of the feature map are documents built from an arbitrary combination of words. This idea is based on Random Features approximation \cite{rahimi2008random} that suggests that a randomized feature map is useful for approximating a shift-invariant kernel.

Our work can be viewed as a combination of the above ideas: we use distance estimates to create interpretable, structured representations of entities with respect to their contexts. The contextual dimension consists of features that are built from an arbitrary combination of discrete labels. 
Our work most importantly differs in the following manners: (1) We use end-to-end deep neural network training to include rich image features when building representations; (2) Information from different modalities (visual and language) can be combined.

\section{CoDiR: Method}

We first define some notions that are useful to understand the method, which we name Composable Distance-based Representation learning (CoDiR).

\subsubsection{Setup and notations}

Given a dataset with data samples  $\rx  \sim \pdata $, with non-exclusive class labels $c_i, i \in \{1,...,n_c\}$ which in this work 
are visual object labels (\textit{e.g.}, dog, ball, ...). Image instances $s$ are fed through a (convolutional) neural network $N$. The outputs of $N$ will serve to build templates $\mT_{i,:} \in \R^{n_e}$ and instance representations $\mD \in \R^{n_c\times n_e}$ with $n_e$ a hyperparameter denoting the amount of environments. Each environment will be defined with the use of discrete environment labels $l_k, k \in \{1,...,n_l\}$, for which we experiment with two types: (1) the same visual object labels as used for the class labels (such that $n_l = n_c$) and (2) image caption keywords from the set of the $n_l$ most common nouns, adjectives or verbs in the sentence descriptions in the dataset. We will refer to the first as \textbf{`CoDiR (class)'} and the latter as \textbf{`CoDiR (capt)'}.

$\1_{c_i}$ is shorthand for the indicator function $\1_{c_i}(x) = 1$ if $x\in C_i$, $0$ otherwise, with $C_i$ the set of images with label $c_i$.  Similarly we denote $\1_{l_k}$. 
Each element $\emD_{i,j}$ is a distance estimate between distributions $p_{c_i}$ and $p_{e_j}$. 
$p_{c_i}$ is shorthand for $p(\rx=x,\rx \in C_i)$.
Informally, $p_{c_i}$ is the joint distribution modeling the data distribution and class membership $c_i$. Similarly, $p_{e_j}$ is shorthand for $p({\rx=x,\rx \in E_j})$, where $E_j=\cup_{m=1}^{\rr_j}L_m^{(j)}$ with $L_m^{(j)}$ the set of images with label $\rl_m^{(j)}$.   
$\rr_j \sim U[1,R] \in \N$ where $R$ is a hyperparameter indicating the maximum amount of labels per environment. For $\rl^{(j)}_m$ with  $m\in\{1,...,\rr_j\}$, labels $l_k$ are sampled uniformly without replacement from the set of all 
labels. For each environment $e_j$, the parameters $\rr_j$ and $\rl^{(j)}_m$ are fixed once before training.

\subsubsection{Contextual distance}
We propose to represent each image as a 2D feature map that 
relates distributions of classes to environments. A suitable metric should be able to deal with neural network training as well as potentially overlapping distributions. A natural candidate is to use the Wasserstein distance \cite{arjovsky2017wasserstein}, which can be understood as the minimal amount of effort that is required to move the mass from one probability distribution to another. A key advantage of using a Wasserstein-based distance function is that the critic can be encouraged to maximize the distance between two distributions, whereas metrics based on Kullback-Leibler (KL) divergence are not well defined if the distributions have a negligible intersection \cite{arjovsky2017wasserstein}. In comparison to other neural network-based distance metrics, the Fisher IPM  provides particularly stable estimates and has the advantage that any neural network can be used as $f$ as long as the last layer is a linear, dense layer \cite{mroueh2017fisher}.
The Fisher GAN formulation bounds $\mathcal{F}$, the set of measurable, symmetric and bounded real valued functions, by construction, that is, by defining a data dependent constraint on its second order moments. The IPM is given by: 
\begin{equation}
	dF_{\mathcal{F}}(p_{e_j},p_{c_i})=\sup_{f_{i,j} \in \mathcal{F} }\frac{ \underset{x\sim p_{e_j}}{\mathbb{E}}[ f_{i,j}(x)] - \underset{x\sim p_{c_i}}{\mathbb{E}}[f_{i,j}(x)]}{ \sqrt{\nicefrac{1}{2}\mathbb{E}_{x\sim p_{e_j} }f_{i,j}^2(x)+\nicefrac{1}{2}\mathbb{E}_{x \sim p_{c_i}} f_{i,j}^2(x)} }
	\label{eq:FisherIPM}
\end{equation}

In practice, the Fisher IPM is estimated with neural network training where the numerator in equation \ref{eq:FisherIPM} is maximized while the denominator is expressed as a constraint, enforced with a Lagrange multiplier. 
While the Fisher IPM is an estimate of the chi-squared distance, the numerator can be viewed as a bounded estimate of the inter-class distance, closely related to the Wasserstein distance \cite{mroueh2017fisher}.  From now on, we denote this approximation of the inter-class distance as the `distance'. During our training, critics $f_{i,j}$ are trained from input images to maximize the Fisher IPM 
for distributions $p_{c_i}$ and $p_{e_j}$, $\forall i\in\{1,...,n_c\},\forall j\in\{1,...,n_e\}$. The numerator then gives the distance between $p_{c_i}$ and $p_{e_j}$. 
We denote $\mT \in \R^{n_c \times n_e}$, with $\emT_{i,j} =\underset{x\sim p_{{e}_j,train}}{\mathbb{E}}[f_{i,j}(x)] - \underset{x\sim p_{c_i,train}}{\mathbb{E}}[f_{i,j}(x)]$, \textit{i.e.}, the evaluation of the estimated distances over the training set. Intuitively, one can see why a matrix $\mT$ with co-occurrence data contains useful information. A subset of images containing `cats', for example, will more closely resemble a subset containing `dogs' and `fur' than one containing `forks' and `tables'.

\subsubsection{Template and instance representations}\label{templates}

As the template representation for class $c_i$, we simply use the corresponding row of the learned distance matrix: $\mT_{i,:}$. Each element $\emT_{i,j}$ gives an average distance estimate for how a class $c_i$ relates to environment $e_j$, where smaller values indicate that class and environment are similar or even (partially) overlap. For the instance representation for an input $s$ we then propose to use $\mD \in \R^{n_c\times n_e}$ with elements given by equation \ref{eq:v_i}: 
\begin{equation}
    \emD_{i,j}^{(s)} =\underset{x \sim p_{{e}_j,train}}{\mathbb{E}}[f_{i,j}(x)] - f_{i,j}(s)
    \label{eq:v_i}
\end{equation} 
where $f_{i,j}(s)$ is simply the output of critic $f_{i,j}$ for the instance $s$.
The result is that for an input $s$ with class label $c_i$, $\mD_{i,:}^{(s)}$ is correlated to $\mT_{i,:}$ as its distance estimates with respect to all different environments should be similar.  
Therefore, the cosine similarity between vector $\mD_{i,:}^{(s)}$ and the template $\mT_{i,:}$ will be large for input samples from class $i$, and small otherwise. 

Such templates can be evaluated, for example, in multi-label classification tasks (see section \ref{exp}). Finding the classes for an image is then simply calculated by computing whether $\forall c_i$, $cos(\mD_{i,:}^{(s)},\mT_{i,:}) > t_{c_i}$ with $t_{c_i}$ a threshold (the level of which is determined during training).
From here on we will use a shorthand notation $\mD^{(s)} \subset c_i$ to denote $cos(\mD_{i,:}^{(s)},\mT_{i,:}) > t_{c_i}$, and $\mD^{(s)} \not\subset c_i$ otherwise.

\subsubsection{Implementation}
Training $n_c \times n_e$ critics is not feasible in practice, so we pass input images through a common neural network for which the classification layer is replaced by 
$n_c \times n_e$ single layer neural networks, the outputs of which constitute $f_{i,j}$ (see Fig. \ref{fig:templates}). 
During training, any given mini-batch will contain inputs with many different $c_i$ and $e_j$. 
To maximize equation \ref{eq:FisherIPM} efficiently, instead of feeding a separate batch for the samples of $x \sim p_{c_i}$  and $x \sim p_{e_j}$, we use the same mini-batch. Additionally, instead of directly sampling $x \sim p_{c_i}$ we multiply each output $f_{i,j}$ with a mask $\mM^c_{i,j}$ where $\mM^c_{i,j} = \1_{c_i}$.
Similarly, for $x \sim p_{e_j}$ we multiply each output $f_{i,j}$ with a mask $\mM^e_{i,j}$ where $\mM^e_{i,j} =  \sum_{m=1}^{\rr_j}\1_{\rl^{(j)}_m}$. The result is that instances then are weighted according to their label prevalence as required.
From these quantities, the Fisher IPM can be calculated and optimized. Algorithm \ref{alg:emd} explains all the above in detail.\footnote{The code will be made available upon acceptance.} When comparing to similar neural network-based methods, the last layer imposes a slightly larger memory footprint ($O(n^2)$ vs $O(n)$) but training time is comparable as they have the same amount of layers.  
After training completes we perform one additional pass through the training set where we use 2/3rd of the samples to calculate the templates and the remaining 1/3rd to set the thresholds for classification.\footnote{All models are trained on a single 12Gb gpu.}

\begin{algorithm*}[t]
	\caption{Algorithm of the training process. For matrices and tensors, $\times$ refers to matrix multiplication and $*$ refers to element-wise multiplication.}
	\label{alg:emd}
	\begin{algorithmic}
	    \STATE Inputs: images $s$, class labels $c$, environment labels $l$
		\STATE $\forall j \in \{1,...,n_e\}$: $\rr_j \sim U[1,R] \in \N \land  \forall m \in \{1,...,r_j\}$, $\rl_m^{(j)} \sim U[1,n_l]$ 
		\STATE Create $\mV \in \N^{n_l \times n_e}$ which has value $1$ for each uniformly selected label, $0$ otherwise.  
		\STATE Init $\mlambda =0 \in \R^{n_c \times n_e}$ $\land$ Init weights in neural network $N$
		\\ 
		\WHILE{Training}
		\STATE Sample a mini-batch $b$, with batch size $n_b$, containing images $s$ and binary class labels $\mC_b \in \N^{n_b \times n_c}$ and binary environment labels $\mL_b \in \N^{n_b \times n_l}$.
		\\
		\STATE \textbf{Create masks}
		\STATE Expand $\mC_b$ 
		into $\mM^c \in \N^{n_b \times n_c \times n_e}$, \textit{s.t.} ${\mM^c}_{k,i,:} = \1_{c_i}(s_k)$ for the $k$-th sample $s_k$. 
		\\
		\STATE Multiply $\mL_b$ and $\mV$, then expand the result 
		into $\mM^e\in \N^{n_b \times n_c \times n_e}$,  \textit{s.t.} ${\mM^e}_{k,:,j} = \sum_{m=1}^{\rr_j}\1_{\rl^{(j)}_m}(s_k)$ for the $k$-th sample $s_k$. 
		\\
		\STATE \textbf{Calculate the FISHER GAN loss}
		\STATE Propagate $b$ through $N$ to obtain $\mO_{f}\in \R^{n_c \times n_e}$ containing all outputs $f_{i,j}$.
		\\
		\STATE Apply masks to $N$'s outputs: $\mO_E = \mO_{f} * \mM^e$ and $\mO_C = \mO_{f} * \mM^c$.  
		\\
		\STATE $E_{fE} = mean(\mO_E, dim=0)$ 
		\STATE $E_{fEs} = mean(\mO_E*\mO_E, dim=0)$
		\STATE $E_{fC} = mean(\mO_C, dim=0)$ 
		\STATE $E_{fCs} = mean(\mO_C*\mO_C, dim=0)$
		\STATE $constraint = 1 - (0.5 * E_{fEs} + 0.5 * E_{fCs})$
		\\
		\STATE Minimize $loss = -sum(
		E_{fE} - E_{fC} + \mlambda * constraint - \rho / 2 * constraint^2)$
		\ENDWHILE
		
	\end{algorithmic}
	
\end{algorithm*}

\subsubsection{(De-)composing representations} \label{composing}

As the CoDiR representations have a clear structure 
, a Singular Value Decomposition of $\mD$: $\mD = \mU\mS\mV$ can be performed, such that the rows of $\mU$ and the columns of $\mV$ can be interpreted as the corresponding factors  as contributed by the $c_i$ and ${e}_j$ respectively. This leads to two applications: 
(1) Composition: by modifying the elements of $\mU$, one can easily obtain $\tilde{\mU}$ with modified information content. By building a new representation $\tilde{\mD}$ from $\tilde{\mU}$, $\mS$ and $\mV$, one thus obtains a similar representation to the original but with modified class membership. This will be further explained in this section. (2) Compression: The spectral norm for instance representations is large 
with a non-flat spectrum. One can thus compress the representations substantially by retaining only the first $k$ eigenvectors of $U$ and $V$, thus creating representations in a lower $k$ dimensional space (rank $k$) without significant loss of classification accuracy. If $k=1$, the new representations are  $(91+300)/(91*300) = 1.4\%$ the size of the original representations. We call this method C-CoDiR($k$). 

Let us consider in detail how to achieve composition. To keep things simple, we only discuss the case for `CoDiR (capt)'. Given an image $s$ for which $\mD^{(s)} \subset c_+$ and $\mD^{(s)} \not\subset c_-$ 
The goal is now to modify $\mD^{(s)}$ such that it represents an image $\tilde{s}$ for which 
$\mD^{(\tilde{s})} \not\subset c_+$ and $\mD^{(\tilde{s})} \subset c_-$ while preserving the contextual information in the environments of $\mD^{(s)}$.  As an example, for a $\mD^{(s)}$ of an image where 
$\mD^{(s)} \subset c_{dog}$ and the discrete labels from which the environments are built indicate labels such as $playing$, $ball$ and $grass$. The goal would be to modify the representation into $\mD^{(\tilde{s})}$ (such that, for example, 
$\mD^{(\tilde{s})} \subset c_{cat}$ and $\mD^{(\tilde{s})} \not\subset c_{dog}$) and to not modify the information in the environments. 

To achieve this, consider that by increasing the value of $\mU_{c_+,:}$, one can increase the distance estimate with respect to class $c_+$, thus expressing that 
$\mD^{(s)} \not\subset c_+$. 
Practically, one can set the values of $\tilde{\mU}_{c_+,:}$ to the mean of all rows in U corresponding to the classes $\bar{c}$ for which $\mD^{(s)} \not\subset \bar{c}$.   
The opposite can be done for class $c_-$, \textit{i.e.}, one can decrease the value of $\mU_{c_-,:}$ such that 
$\mD^{(\tilde{s})} \subset c_-$. To set the values of $\tilde{\mU}_{c_-,:}$, one can perform a SVD on the matrix composed of all $n_c$ template representations $\mT$, thus obtaining $\mU_\mT\mS_\mT\mV_\mT$. As the templates by definition contain estimated distances for samples of all classes, it is then easy to see that by setting $\tilde{\mU}_{c_-,:} = \mU_{\mT_{c_-,:}}$ we express that 
$\mD^{(\tilde{s})} \subset c_-$ as desired. 
A valid representation can then be reconstructed with the outer product $\mD^{(\tilde{s})} = \underset{k}{\sum} \sigma_k \tilde{\mU}_{:,k} \otimes \mV_{k,:}^{\top} $ where $\sigma_k$ are the eigenvalues of $\mD^{(s)}$. In the next section this is illustrated by retrieving images after modifying the representations.

\section{Experiments} \label{exp}

We show how CoDiR compares to a (binary) cross-entropy baseline for multi-label image classification. Additionally, CoDiR's qualities related to (de)compositions, compression and rank are examined.

\subsection{Setup.}
\begin{table*}[t]
	\caption{F1 scores, precision (PREC) and recall (REC) for different models for the multi-label classification task. $\sigma$ is the standard deviation of the F1 score over three runs. 
	All results are the average of three runs. }
	\label{table:ours_vs_bxent}
	\begin{center}
		\begin{tabular}{lllllllll}
			\multicolumn{1}{c}{\bf MODEL} &\multicolumn{1}{c}{\bf METHOD}  &\multicolumn{1}{c}{\bf $n_e$}  &\multicolumn{1}{c}{\bf $n_l$}  &\multicolumn{1}{c}{\bf $R$} &\multicolumn{1}{c}{\bf F1} &\multicolumn{1}{c}{\bf PREC} &\multicolumn{1}{c}{\bf REC} &\multicolumn{1}{c}{\bf $\sigma$}
			\\ \hline 
			ResNet-18&BXENT (single) & - & - & - & 0.566 &  0.579 & 0.614 & $3.6e^{-3}$ \\
			ResNet-18&\bf CoDiR (class)  & 300 & 91 & 40 &  \bf0.601 & 0.650 &  0.613 & $8.0e^{-3}$ \\  
			\hline
			ResNet-101&BXENT (single) & - & - & - & 0.570 &0.582 &0.623 &  $1.3e^{-2}$\\ 
			ResNet-101&\bf CoDiR (class)  & 300 & 91 & 40 & \bf 0.627 & 0.664 &0.648 &  $2.5e^{-3}$\\
			\hline
			Inception-v3&BXENT (single)  & - & - & - & \bf 0.638 &0.663 &0.669 &  $5.4e^{-3}$\\
			Inception-v3&\bf CoDiR (class) & 300 & 91 & 40 & 0.617 &0.648 &0.646 &  $4.7e^{-3}$\\
			
			\hline
			\hline
			
			ResNet-18&BXENT (joint)  & - & 300 & - & 0.611 &  0.631 & 0.654 & $1.1e^{-3}$\\ 
			ResNet-18&BXENT (joint)  & - & 1000 & - & 0.614 &	0.637 &	0.653 & $9.3e^{-3}$\\ 
			
			ResNet-18&\bf CoDiR (capt)  & 300 & 300 & 40 &  0.629 & 0.680 &  0.641 &  $2.7e^{-3}$ \\  
			ResNet-18&\bf CoDiR (capt)  & 1000 & 1000 & 100 & \bf 0.638 & 0.686 &  0.651 &  $1.9e^{-3}$ \\
			
			\hline
			
			ResNet-101&BXENT (joint) & - & 300 & - & 0.598 &0.619 &0.640 &  $1.1e^{-2}$\\ 
			ResNet-101&BXENT (joint) & - & 1000 & - &  0.592 &0.611 &0.638 &  $7.0e^{-3}$\\ 
			ResNet-101&\bf CoDiR (capt)  & 300 & 300 & 40 &  0.645 &0.696 &0.655 &  $2.8e^{-2}$\\
			ResNet-101&\bf CoDiR (capt)  & 1000 & 1000 & 100 &\bf 0.657 &0.702 &0.666 &  $1.3e^{-2}$\\
			
			\hline
			
			Inception-v3&BXENT (joint) & - & 300 & - & 0.644 &	0.671 &	0.675 &  $1.5e^{-2}$\\ 
			Inception-v3&BXENT (joint) & - & 1000 & - & 0.63 &	0.655 &	0.663 &  $3.0e^{-2}$\\
			
			Inception-v3&\bf CoDiR (capt) &300 & 300 & 40 &  0.660 &0.699 &0.675 &  $1.9e^{-3}$ \\
			Inception-v3&\bf CoDiR (capt) &1000 & 1000 & 100 & \bf 0.661 &0.700 &  0.676 &  $6.5e^{-3}$ \\

		\end{tabular}
	\end{center}
\end{table*}
The experiments are performed on the COCO dataset \cite{lin2014microsoft} which contains multiple labels and descriptive captions for each image. We use the 2014 train/val splits of this dataset as these sets contain the necessary labels for our experiment, where we split the validation set into two equal, arbitrary parts to have a validation and test set for the classification task. 
We set $n_c = 91$, \textit{i.e.}, we use all available 91 class labels (which includes 11 supercategories that contain other labels, \textit{e.g.}, `animal' is the supercategory for `zebra' and `cat'). 
An image can contain more than one class label. To construct environments we use either the class labels, CoDiR (class), or the captions, CoDiR (capt). For the latter, a vocabulary is built of the $n_l$ most frequently occurring adjectives, nouns and verbs. For each image, each of the $n_l$ labels is then assigned if the corresponding vocabulary word occurs in any of the captions. 
For the retrieval experiment we select a set of 400 images from the test set and construct their queries.\footnote{All dataset splits and queries will be published upon acceptance}
All images are randomly cropped and rescaled to $224 \times 224$ pixels. We use three types of recent state-of-the-art classification models to compare performance: ResNet-18, ResNet-101 \cite{he2016deep} and Inception-v3 \cite{szegedy2016rethinking}. For all runs, an Adam optimizer is used with learning rate $5.e-3$. $\rho$ for the Fisher IPM loss is set to $1e^{-6}$. Parameters are found empirically based on performance on the validation set.

\subsection{Results}\label{exp:results}

\textbf{Multi-label image classification.}
In this experiment the objects in the image are recognized. 
For each experiment images are fed through a neural network where the only difference between the baseline and our approach is the last layer. For the baseline, which we call `BXENT', the classification model is trained with a binary cross-entropy loss over the outputs and optimal decision thresholds are selected based on the validation set F1 score. For CoDiR, classification is performed on the learned representations as explained in section \ref{templates}. 
We then conduct two types of experiments: (1) \textbf{BXENT (single)} vs \textbf{CoDiR (class)}: An experiment where only class labels are used. For BXENT (single), classification is performed on the output with dimension $n_c$. For CoDiR (class), environments are built with class labels, such that $n_l$ = $n_c$. (2) \textbf{BXENT (joint)} vs \textbf{CoDiR (capt)}: An experiment where $n_l$ additional contextual labels from image captions are used. The total amount of labels is $n_c + n_l$. For BXENT (joint) this means joint classification is performed on all $n_c+n_l$ outputs. For CoDiR (capt), there are $n_c$ classes whereas environments are built with the selected $n_l$ caption words. For all models, scores are computed over the $n_c$ class labels.

With the same underlying architecture, table \ref{table:ours_vs_bxent} shows that the CoDiR method compares favorably to the baselines in terms of F1 score.\footnote{Multi-label scores as defined by \cite{sorower2010literature}.} When adding more detailed contextual information in the environments, as is the case for CoDiR (capt), our model outperforms the baseline in all cases.\footnote{For reference: a k-Nearest Neighbors ($k=3$) on pre-trained ImageNet features of a ResNet-18 achieves a F1 of 0.221.} 

The performance of CoDiR depends on the parameters $n_e$ and $R$. To measure this influence the multi-label classification task is performed for different $n_e$ values. 
Increasing $n_e$ or the amount of environments (\textit{i.e.}, the amount of columns of the CoDiR representation) leads in general to better performance, although it plateaus after a certain level. For $R$ an optimal value can also be found empirically between $0$ and $n_l$. The reason is that combining a large amount of 
labels in any environment 
creates a unique subset to compare samples with. When $R$ is too large, however, subsets with unique features are no longer created and performance deteriorates. 
Also, even when $n_e$ and $R$ are small, the outcome is not sensitive with regard to the choice of environments, suggesting that the amount and diversity are more important than the composition of the environments.

\begin{figure}[t]
	\begin{center}
		\includegraphics[width=0.99\linewidth]{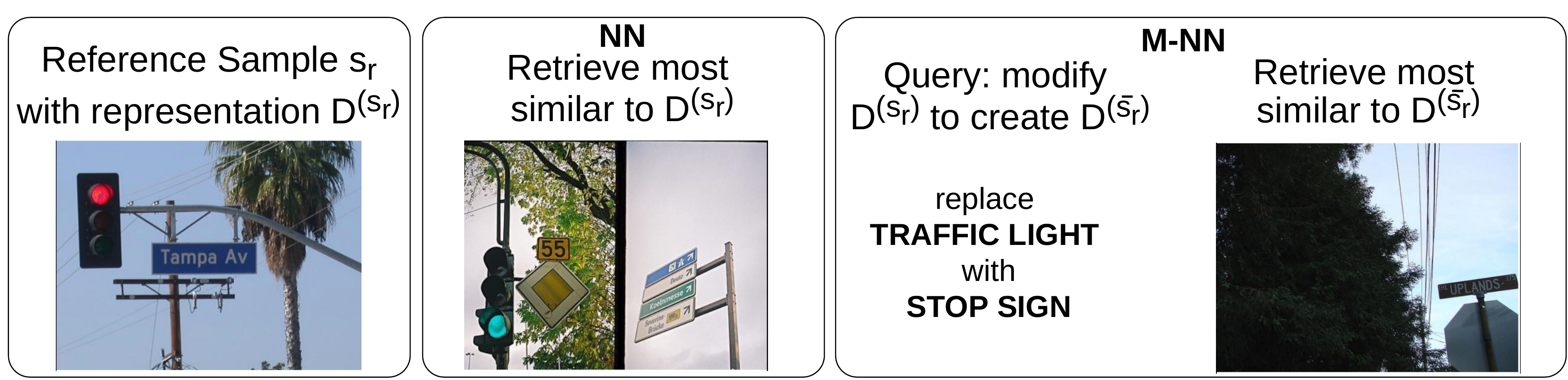}
	\end{center}
	\caption[Retrieval example]{Example of a retrieval result for both NN and M-NN. For NN, based on the representation $\mD^{(s_r)}$, the most similar instance is retrieved. For M-NN $\mD^{({s_r})}$ is modified into $\mD^{(\bar{s_r})}$ before retrieving the most similar instance. }
	\label{fig:retrieval}
\end{figure}

\textbf{Retrieval.} 
The experiments here 
are designed to show interpretability, composability and compressibility of the 
CoDiR representations. All models and baselines in these sections are 
pre-trained on the classification task above. 
We perform two types of retrieval experiments: (1) NN: the most similar sample to a reference sample is retrieved; (2) M-NN: a sample is retrieved with modified class membership while contextual information in the environments is retained. Specifically:
``Given an input $s_r$ that belongs to class $c_{+}$ but not $c_{-}$, retrieve the instance in the dataset that is most similar to $s_r$ that belongs to $c_{-}$ and not $c_{+}$'', where  $c_{+}$ and $c_{-}$ are class labels (see Fig. \ref{fig:retrieval}). We will show that CoDiR is well suited for such a task, as its structure can be exploited to create modified representations $\mD^{(\bar{s_r})}$ through decomposition as explained in section \ref{composing}.

\begin{table}
    \caption{Methods are used in combination with three different base models: ResNet-18/ ResNet-101/ Inception-v3).
    All results are the average of three runs.}
    \begin{subtable}[t]{0.57\textwidth}

		\begin{tabular}{l|ccc}
			\multicolumn{1}{c}{\bf Method }  &\multicolumn{1}{c}{\bf NN}  &\multicolumn{1}{c}{\bf M-NN } & 	\\
			&\multicolumn{1}{c}{\bf  F1}  &\multicolumn{1}{c}{\bf  PREC}  & \multicolumn{1}{c}{\bf  F1$\%$}  
			\\\hline 
			SEM(single)  &.64/.66/.70   & .53/.55/.55&93/87/89  \\
			SEM(joint)  & .71/.70/.73  & .29/.28/.31  & 97/100/96 \\
			\hline
			CNN(joint)   &.71/.70/.70   & .37/.26/.33&92/90/92   \\
			 CM  &.72/.74/.74   & .19/.15/.18   & 100/100/100\\
			 \hline
			CoDiR& .70/.72/.72	 &	.30/.30/.27   & 97/97/95 \\ 
			C-CoDiR(5) &.70/.72/.72 &.30/.29/.26  & 97/94/93\\ 
		\end{tabular}
		\caption{For the NN and M-NN retrieval, the F1 score of class labels and the precision (PREC) of the modified labels are shown for the first retrieved sample.  The proportion of the F1 score of M-NN over NN for the caption words is shown as F1$\%$. }
		\label{table:retrieval}

    \end{subtable}
    \hspace{\fill}
    \begin{subtable}[t]{0.41\textwidth}
       
            \begin{tabular}{l|c}
                \multicolumn{1}{c}{\bf Method }   & \multicolumn{1}{c}{\bf  F1}  
                \\
                \hline 
                SEM (single)  & 0.00/0.00/0.00 \\    		
                CoDiR (class)  &  0.10/0.06/0.07 \\ 
                C-CoDiR(5)(class)  & 0.06/0.08/0.09 \\ 
                \hline
                SEM (joint)  & 0.00/0.10/0.00 \\    
                CoDiR (capt) & 0.08/0.15/0.20 \\ 
                C-CoDiR(5)(capt) & 0.10/0.14/0.19    \\ 
            \end{tabular}
            \caption{F1 score for a simple logistic regression on pre-trained representations to classify a previously unseen label ("panting dogs"). For the last three models, $n_l=300$. }
            \label{table:panting_dog}

    \end{subtable}
\end{table}

This task is evaluated as shown in table \ref{table:retrieval} where the goal is to achieve a good combination of M-NN PREC and F1\% (for the latter, higher percentages are better).
We use the highly structured sigmoid outputs of the BXENT (single) and BXENT (joint) models as baselines, denoted as \textbf{SEM (single)} and \textbf{SEM (joint)} respectively. With SEM (joint) it is possible to directly modify class labels while maintaining all other information. It is thus a `best-case scenario'-baseline for which one can strive, as it combines a good M-NN precision and F1\% score. SEM (single) on the other hand only contains class information and thus presents a best-case scenario for the M-NN precision score yet a worst-case scenario for the F1\% score. Additionally we compare with a simple baseline consisting of \textbf{CNN} features from the penultimate layer of the BXENT (joint) models with $n_l=300$. We also use those features in a \textbf{Correlation Matching (CM)} baseline, that combines different modalities (CNN features and word caption labels) into the same representation space \cite{rasiwasia2010new}. The representations of these baseline models cannot be composed directly. In order to compare them to the `M-NN' method, therefore, we define templates as the average feature vector for a particular class. We then modify the representation for a sample $s$ by subtracting the template of $c_+$ and adding the template of $c_-$. All representations except SEM (single) are built from the BXENT (joint) models with $n_l = 300$. For CoDiR they are built from CoDiR (capt) with $n_l = 300$. 

For all baselines similarity is computed with the cosine similarity, whereas for CoDiR we exploit its structure as:
$similarity =$ $mean\_cos{(D^{(\bar{s_r})},D^{(s)})}$ over all classes $c$ for which $cos(D_{c,:}^{(\bar{s_r})}, T_c) > 0.75 \times t_c$ . 
Here, notations are taken from section \ref{composing} and $\mD^{(\bar{s_r})}$ is the modified representation of the reference sample. $mean\_cos(D^{(\bar{s_r})},D^{(s)})$ is the mean cosine similarity between $\mD^{(\bar{s_r})}$ and $\mD^{(s)}$  with the mean calculated over class dimensions. The similarity is thus calculated over class dimensions where classes with low relevance, \textit{i.e.}, those that have a low similarity with the templates, are not taken into account. 

The advantages of the composability of the representations can be seen in table \ref{table:retrieval} where CoDiR (capt) has comparable performance to the fully semantic SEM (joint) representations. CNN (joint) manages to obtain a decent M-NN precision score, thus changing class information well, but at the cost of losing contextual information (low F1\%), performing almost as poorly as SEM (single). Whereas CM performs well on the NN task, it doesn't change the class information accurately and thus (inadvertently) retains most contextual information. 
		
\textbf{Rank.} 
While the previous section shows that the structure of CoDiR representations provides access to semantic information derived from the labels on which they were trained, we hypothesize that the representations contain additional information beyond those labels, reflecting local, continuous features in the images. To investigate this hypothesis, we perform an experiment, similar to \cite{yang2018breaking}, to determine the rank of a matrix composed of 1000 instance representations of the test set. To maintain stability we take only the first 3 rows (corresponding to 3 classes) and all 300 environments of each representation. Each of these is flattened into a 1D vector of size 900 to construct a matrix of size 1000*900. Small singular values are thresholded as set by \cite{press2007numerical}. The used model is the CoDiR (capt) ResNet-18 model with $n_l=300$.
We obtain a rank of 499, which exceeds the amount of class and environment labels (3+300) within, suggesting that the representations contain additional structure beyond the original labels. 

 The representations can thus be compressed. Table \ref{table:retrieval} shows that C-CoDiR with $k=5$, denoted as \textbf{C-CoDiR(5)}, approaches CoDiR's performance across all defined retrieval tasks. 
 To show that the CoDiR representations contain information beyond the pre-trained labels, we also use cross-validation to perform a binary classification task with a simple logistic regression. A subset of 400 images of dogs is taken from the validation and test sets, of which 24 and 17 respectively are positive samples of the previously unseen label:
 \textit{panting dogs}. 
 The outcome in table \ref{table:panting_dog} shows that CoDiR and C-CoDiR(5) representations outperform the purely semantic representations of the SEM model, which shows that the additional continuous information is valuable.

\section{Conclusion} 

CoDiR is a novel deep learning method to learn representations that can combine different modalities. The instance representations are obtained from images with a convolutional neural network and are structured along class and environment dimensions. Templates are derived from the instance representations that generalize the class-specific information. In a classification task it is shown that this generalization improves as richer contextual information is added to the environments. When environments are built with labels from image captions, the CoDiR representations consistently outperform their respective baselines. The representations are continuous and have a high rank, as demonstrated by their ability to classify a label that was not seen during pre-training with a simple logistic regression. At the same time, they contain a clear structure which allows for a semantic interpretation of the content. It is shown in a retrieval task that the representations can be decomposed, modified and recomposed to reflect the modified information, while conserving existing information. 

CoDiR opens an interesting path for deep learning applications to explore uses of structured representations, similar to how such structured matrices played a central role in many language processing approaches in the past. In zero-shot settings the structure might be exploited, for example, to make compositions of classes and environments that were not seen before. Additionally, further research might explore unsupervised learning or how the method can be applied to other tasks and modalities with alternative building blocks for the environments. While we demonstrate the method with a Wasserstein-based distance, other distance or similarity metrics could be examined in future work.

\section*{Acknowledgements}
This work was partly supported by the FWO and SNSF, grants G078618N and \#176004 as well as an ERC Advanced Grant, \#788506.
	
%
%
%
\bibliographystyle{splncs04}
\bibliography{mybib}

\end{document}